\documentclass[pdflatex,sn-mathphys-num]{sn-jnl}% Math and Physical Sciences Numbered Reference Style 
\usepackage{graphicx}%
\usepackage{multirow}%
\usepackage{amsmath,amssymb,amsfonts}%
\usepackage{amsthm}%
\usepackage{mathrsfs}%
\usepackage[title]{appendix}%
\usepackage{xcolor}%
\usepackage{textcomp}%
\usepackage{manyfoot}%
\usepackage{booktabs}%
\usepackage{algorithm}%
\usepackage{algorithmicx}%
\usepackage{algpseudocode}%
\usepackage{listings}%
\theoremstyle{thmstyleone}%
\theoremstyle{thmstyletwo}%

\theoremstyle{thmstylethree}%

\begin{document}

\title[Article Title]{Deployment Pipeline from Rockpool to Xylo for Edge Computing}


\author[1]{Peng Zhou}\email{zp.pengzhou@gmail.com}

\author[2]{Dylan R. Muir}\email{dylan.muir@synsense.ai}

\affil[1]{LuxiTech, Shenzhen, China}

\affil[1]{SynSense, Zurich, Switzerland}


\abstract{Deploying Spiking Neural Networks (SNNs) on the Xylo neuromorphic chip via the Rockpool framework represents a significant advancement in achieving ultra-low-power consumption and high computational efficiency for edge applications. This paper details a novel deployment pipeline, emphasizing the integration of Rockpool’s capabilities with Xylo’s architecture, and evaluates the system’s performance in terms of energy efficiency and accuracy. The unique advantages of the Xylo chip, including its digital spiking architecture and event-driven processing model, are highlighted to demonstrate its suitability for real-time, power-sensitive applications.}




\maketitle

\section{Introduction}\label{sec1}
Deploying neural networks onto custom Application-Specific Integrated Circuit (ASIC) hardware involves comprehensive integration across all levels of the technology stack, including application design, algorithm development, compilation techniques, and chip architecture. This multi-layered process is fraught with challenges due to the need for precise alignment between high-level network functionality and low-level hardware operations. Here, we present an extensible deployment pipeline utilizing the Rockpool framework and the Xylo neuromorphic chip.

\section{Rockpool}\label{sec2}
Rockpool \cite{rockpool} is a open-source Python package designed specifically for working with dynamical neural network architectures, particularly adept at designing event-driven networks for neuromorphic computing hardware. It offers a robust interface for designing, training, and evaluating recurrent networks that can operate with both continuous-time dynamics and event-driven dynamics. It provides a high-level API compatible with various backends like PyTorch, Jax, and Numpy. Rockpool supports hardware-aware training which is critical for optimizing models for specific neuromorphic hardware such as the Xylo chip.

\section{Xylo}\label{sec3}
The Xylo chip \cite{bos2023sub, bos2024micro} is an ultra-low-power neuromorphic ASIC optimized for the efficient simulation of spiking leaky integrate-and-fire neurons with exponential input synapses. It is highly configurable, supporting individual synaptic and membrane time-constants, thresholds, and biases for each neuron, and can support arbitrary network architectures, including recurrent networks, for up to 1000 neurons. The integration of Rockpool's model design and training capabilities with Xylo's configurable, efficient hardware platform exemplifies a leading-edge approach to deploying energy-efficient neuromorphic computing at the edge.

The following table details the most important hardware specifications of the Xylo chip that should be considered when designing networks for deployment:

\begin{table}[h]
\centering
\begin{tabular}{|l|c|}
\hline
\textbf{Description} & \textbf{Number} \\ \hline
Max. input channels & 16 \\ \hline
Max. input spikes per time step & 15 \\ \hline
Max. hidden LIF neurons & 1000 \\ \hline
Max. hidden neuron spikes per time step & 31 \\ \hline
Max. input synapses per hidden neuron & 2 \\ \hline
Max. alias targets (hidden neurons only) & 1 \\ \hline
Max. output LIF neurons & 8 \\ \hline
Max. output neuron spikes per time step & 1 \\ \hline
Max. input synapses per output neuron & 1 \\ \hline
Weight bit-depth & 8 \\ \hline
Synaptic state bit-depth & 16 \\ \hline
Membrane state bit-depth & 16 \\ \hline
Threshold bit-depth & 16 \\ \hline
Bit-shift decay parameter bit-depth & 4 \\ \hline
Max. bit-shift decay value & 15 \\ \hline
Longest effective time-constant & 32768 \(\cdot\) dt (but subject to linear decay) \\ \hline
\end{tabular}
\caption{Key Specifications of the Xylo™Audio 2 Neuromorphic chip}
\label{tab:xylo-specs}
\end{table}

\section{Deployment Pipeline}\label{sec2}

% The deployment pipeline is developed with rockpool Pytorch backend
The deployment pipeline for Rockpool networks onto the Xylo Hardware Development Kit (HDK) involves a series of structured steps to ensure that the neural network model is optimized for neuromorphic hardware implementation. The flow-chart below summarizes these steps, beginning with network construction in Rockpool and culminating in network simulation on the Xylo HDK.

\begin{figure}[htbp]
\centering
\includegraphics[width=\textwidth]{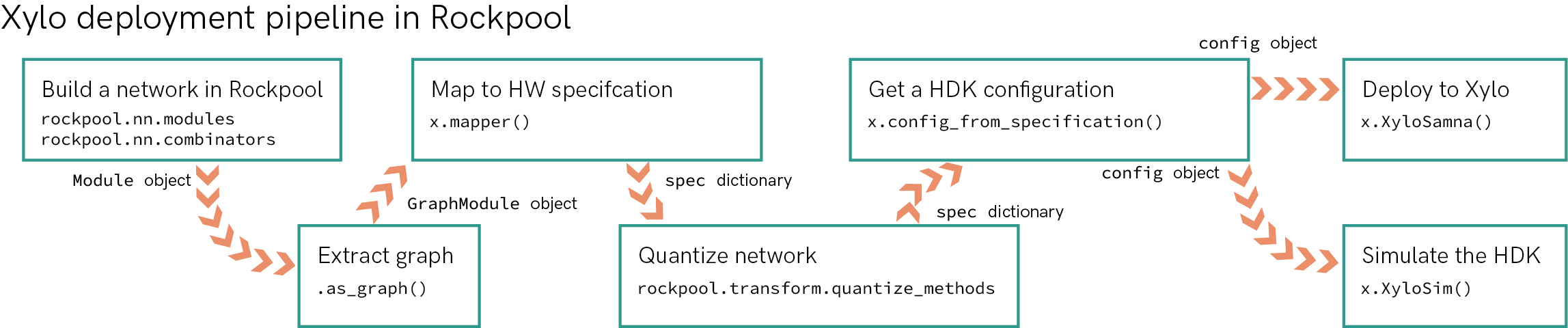}
\caption{Flowchart summarizing the steps from network definition to deployment on the Xylo HDK using Rockpool.\cite{rockpool}}
\label{fig:deployment_pipeline}
\end{figure}

% Each phase of the pipeline is designed to systematically transform the high-level network architecture into a configuration that is compatible with the hardware constraints and capabilities of the Xylo HDK. This process ensures that the final deployed model retains the intended functionality while operating efficiently on neuromorphic hardware.

\subsection{Building a Network in Rockpool}
Utilizing \texttt{rockpool.nn.modules} such as layers \texttt{LIF} and \texttt{Linear}, we can construct neural network architecture. These modules serve as the building blocks for creating diverse and complex neural network configurations tailored to specific tasks and functionalities. Additionally, we use \texttt{rockpool.nn.combinators} like \texttt{Sequential} and \texttt{Residual} to add neural network connections, enhancing the flexibility and capability of the network architecture.

\begin{figure}[htbp]
\centering
\includegraphics[width=1\textwidth]{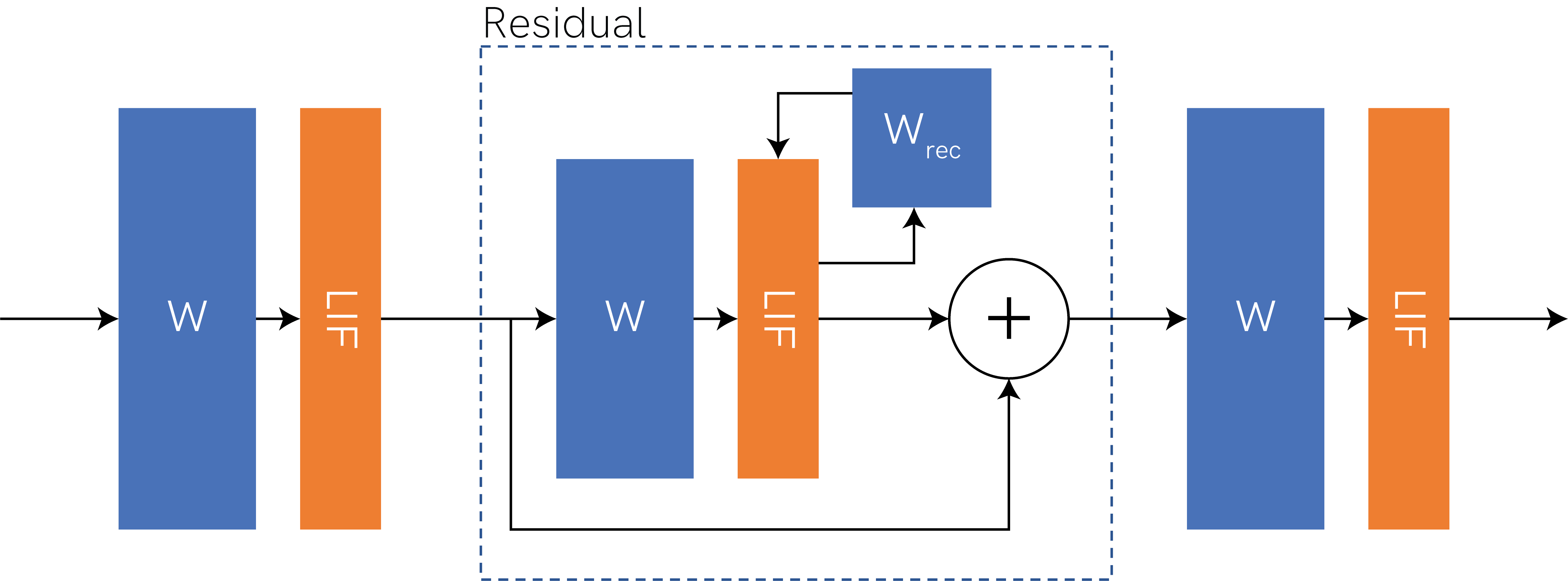}
\caption{Illustration of a neural network configuration using \texttt{Sequential} and \texttt{Residual} combinators in Rockpool, with both feedforward and recurrent weight architectures.\cite{rockpool}}
\label{fig:ResidualNetwork}
\end{figure}

An example of utilizing these tools is shown in Figure \ref{fig:ResidualNetwork}, where the network employs both a \texttt{Residual} architecture as well as feedforward and recurrent blocks, demonstrating how complex architectures can be easily composed.
%This setup is particularly significant in architectures that benefit from residual connections, as they help in learning complex patterns without the vanishing gradient problem by allowing gradients to flow through a shortcut connection.

\subsection{Extracting a computational Graph}
The graph extraction process in Rockpool is a critical step for preparing a neural network model for deployment onto neuromorphic hardware such as the Xylo chip. This process transforms the complex, dynamic structures of a neural network into a static graph format that can be efficiently mapped and deployed onto Neuromorphic hardware.

\subsubsection{GraphModule and GraphNode}
\begin{figure}[htbp]
\centering
\includegraphics[width=0.8\textwidth]{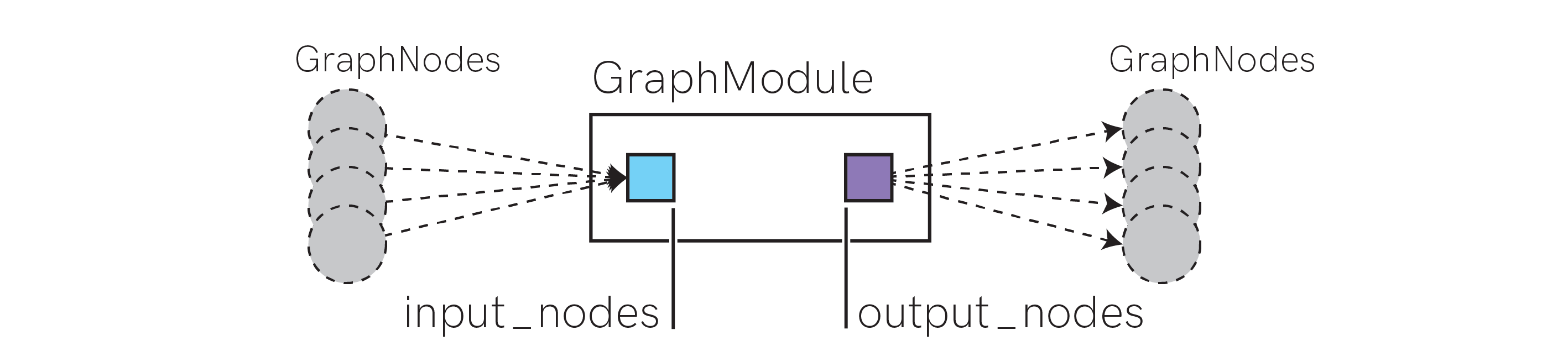}
\caption{Illustration of a \texttt{GraphModule} showing its components and connections.}
\label{fig:GraphModule}
\end{figure}

\begin{figure}[htbp]
\centering
\includegraphics[width=0.8\textwidth]{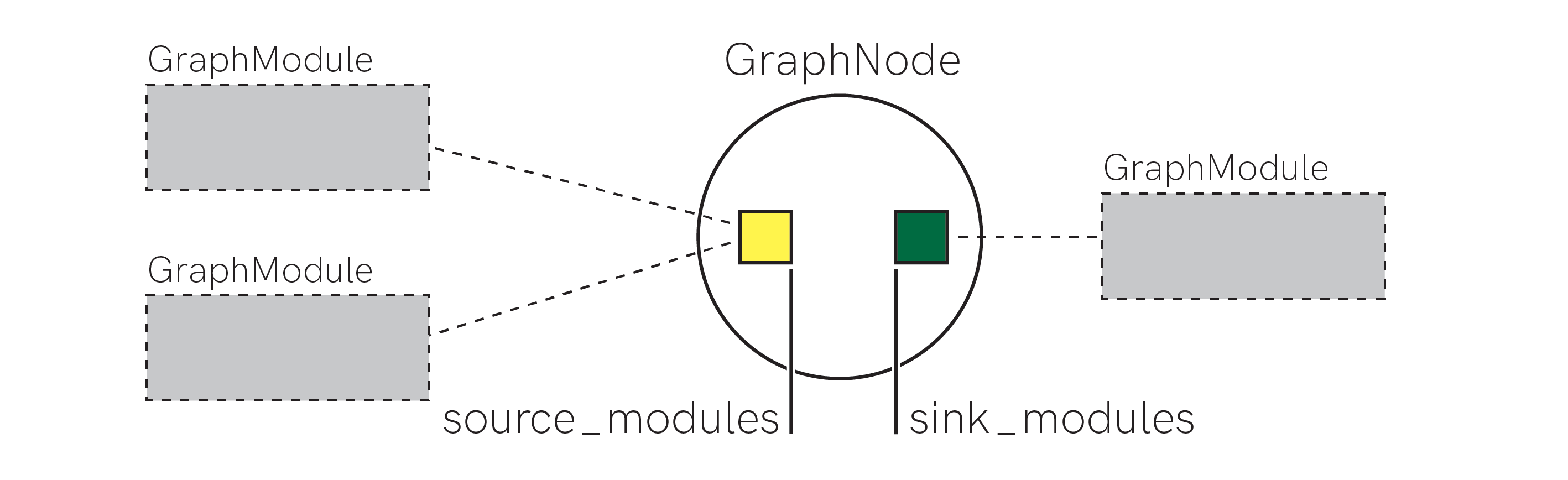}
\caption{Illustration of a \texttt{GraphNode} showing how it acts as a connector within the network.}
\label{fig:GraphNode}
\end{figure}

The core of Rockpool's graph extraction process involves the use of \texttt{GraphModule} and \texttt{GraphNode} classes. A \texttt{GraphModule} acts as a unit of computation and encapsulates the functional aspects of parts of the network, such as layers or neuron groups, as shown in Fig. \ref{fig:GraphModule}. Each \texttt{GraphModule} is linked to input and output through \texttt{GraphNode} objects, which manage the flow of data through the network. The \texttt{GraphNode} essentially serves as the connectivity tissue, directing the input and output to and from \texttt{GraphModule}, as shown in Fig. \ref{fig:GraphNode}.

\subsubsection{Constructing a Graph}
To construct a graph, you begin by defining \texttt{GraphModule}s that represent the different computational components of your network, ranging from simple sets of weights to complex configurations of spiking neurons. Each \texttt{GraphModule}  includes the parameters necessary for defining the functionality of the unit, such as time constants and biases in the case of LIF neuron layers, or sets of weights in the case of linear layers.

Modules can be instantiated by providing existing input and output nodes explicitly or by utilizing the \texttt{\_factory()} method. This method automates the creation of new input and output nodes, ensuring that all necessary attributes and configurations are correctly initialized. When instantiating a \texttt{GraphModule} subclass, additional arguments may be provided to the constructor or the \texttt{\_factory()} method to customize the module according to specific requirements of the neural network being designed.

This setup allows for the precise control and routing of data within the graph, ensuring that each component interacts correctly with others based on the network architecture.

% Each \texttt{GraphModule} must be explicitly connected using \texttt{graph.GraphModule} objects, which are defined to handle the specific input-output relationships necessary for the model's operations. This setup allows for the precise control and routing of data within the graph, ensuring that each component interacts correctly with others based on the network’s architecture.

\subsubsection{Connecting Modules}
\begin{figure}[htbp]
\centering
\includegraphics[width=0.8\textwidth]{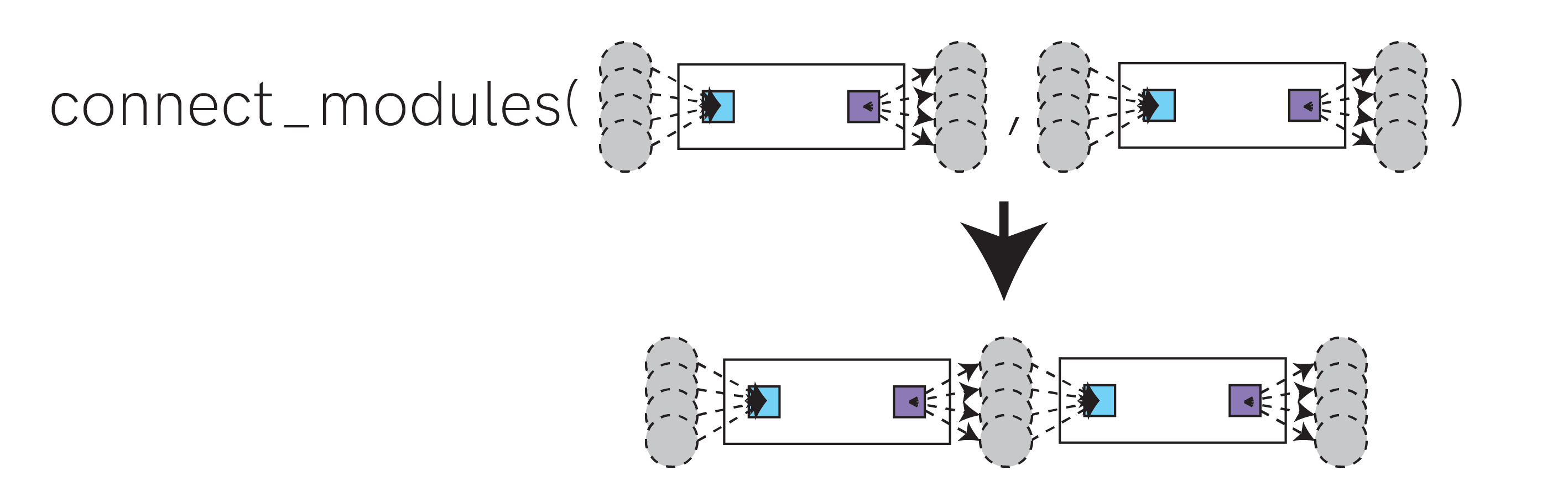}
\caption{Demonstration of connecting modules using \texttt{connect\_modules()} in Rockpool. This function integrates output nodes from one module with the input nodes of another, effectively chaining the computational flow across modules.}
\label{fig:connectModules}
\end{figure}

Connecting these modules requires careful management of \texttt{graph.GraphNode}. The \texttt{graph.utils.connect\_modules()} function in Rockpool is typically used to link output nodes from one module to input nodes of another, establishing a direct pathway for data flow. This function is vital for maintaining the integrity of data transformations across the network. An illustration of how modules are connected using the \texttt{connect\_modules()} function is shown in Fig. \ref{fig:connectModules}.

\subsubsection{Finalizing the Graph}
Once all modules are interconnected, the final graph represents a comprehensive map of the network, detailing how data should move and be processed across different computational units. This graph is then ready to be translated into a format suitable for deployment on hardware like the Xylo chip.

When you have built up a graph, you can conveniently encapsulate complex graphs and subgraphs using the \texttt{graph.GraphHolder} class. This class only requires the \texttt{input\_nodes} and \texttt{output\_nodes} arguments and conceptually holds an entire subgraph linked from those nodes. You can also use the \texttt{as\_GraphHolder()} function to easily encapsulate a \texttt{GraphModule} as a \texttt{GraphHolder}. Graphs can be traversed by iterating over \texttt{GraphModule.input\_nodes}, \texttt{GraphNode.sink\_modules}, etc.

The following Fig. \ref{fig:GraphHolder} illustrates the structure of a \texttt{GraphHolder} encapsulating a network's modules and nodes, highlighting the modular architecture and the flow of data across the system.

\begin{figure}[htbp]
\centering
\includegraphics[width=1\textwidth]{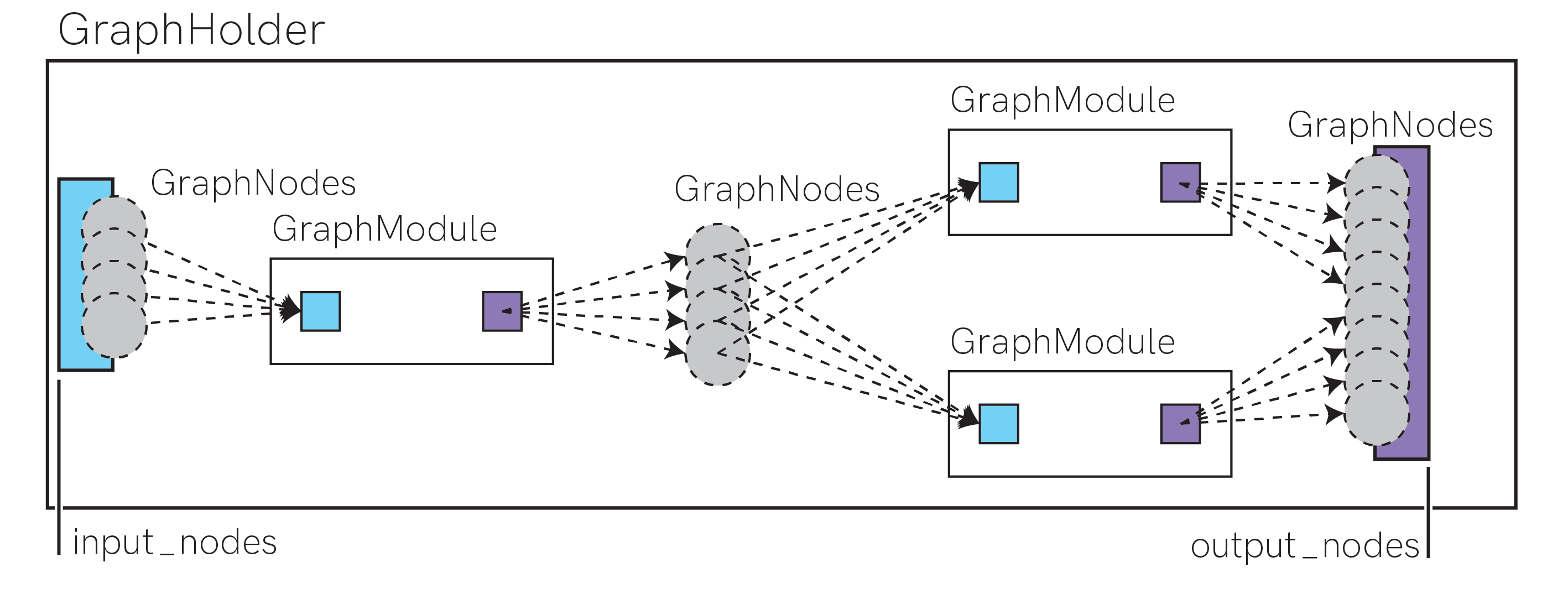}
\caption{Illustration of the \texttt{GraphHolder} encapsulating a network's modules and nodes, demonstrating how data flows and is managed across the network.}
\label{fig:GraphHolder}
\end{figure}

% The graph extraction step is crucial because it translates dynamic, potentially non-linear operations of a neural network into a structured format that neuromorphic hardware can interpret and execute. This ensures that the deployed model operates under the constraints and capabilities of the hardware, optimizing both performance and power efficiency.

\subsection{Map to Hardware Specification}
Mapping the network graph to the Xylo hardware specifications is a crucial step that ensures the network's architecture can be effectively supported by the hardware. This process involves assigning hardware resources to various components of the network. For instance, each neuron within the network must be matched with a hardware neuron on the Xylo chip, and each non-input weight must be associated with a global hidden weight matrix specific to Xylo's architecture. Additionally, neurons in the final layer of the network need to be assigned to output channels to facilitate the correct delivery of output signals.

The Xylo family includes several devices, each with different hardware blocks and capabilities, which are catered to by specific subpackages within \texttt{rockpool.devices.xylo}, named according to the chip ID recognized by your Hardware Development Kit (HDK). Rockpool automates the process of detecting a connected HDK and loading the appropriate support package. This functionality is particularly helpful when setting up the environment for network deployment, as it ensures that the correct device-specific configurations are applied.

If a Xylo HDK is connected, Rockpool identifies it and imports the necessary package for interfacing with the hardware. For users without a Xylo HDK, there is support available for Xylo-Audio 2 through the \texttt{rockpool.devices.xylo.syns61201} package.

Once the appropriate device package is loaded, the \texttt{x.mapper()} function is used to convert the computational graph into a Xylo-compatible specification. This function also allows for specifying parameter types, such as retaining floating-point representations for weights, which is essential for maintaining the precision required for certain computations.

% The mapping process not only ensures that the neural network's design is feasible on the selected hardware but also optimizes the configuration to leverage the unique capabilities of the Xylo hardware, enhancing performance and efficiency.

\subsection{Network Quantization}
The network is quantized using \texttt{rockpool.transform.quantize\_methods} to match the precision requirements of the Xylo chip. The quantization process can be approached using either global or channel-specific methods, each catering to different needs and resulting in various levels of precision and performance efficiency.

\subsubsection{Global Quantization}
Under the global quantization approach, all weights in the network are considered together when scaling and quantizing weights and thresholds. This method treats input and recurrent weights as one group, while output weights are processed separately. The advantage of this method is that it maintains a uniform scale across all parameters, simplifying the quantization process but possibly at the cost of losing some precision where finer control over individual weights might be beneficial.

\begin{figure}[htbp]
\centering
\includegraphics[width=0.4\textwidth]{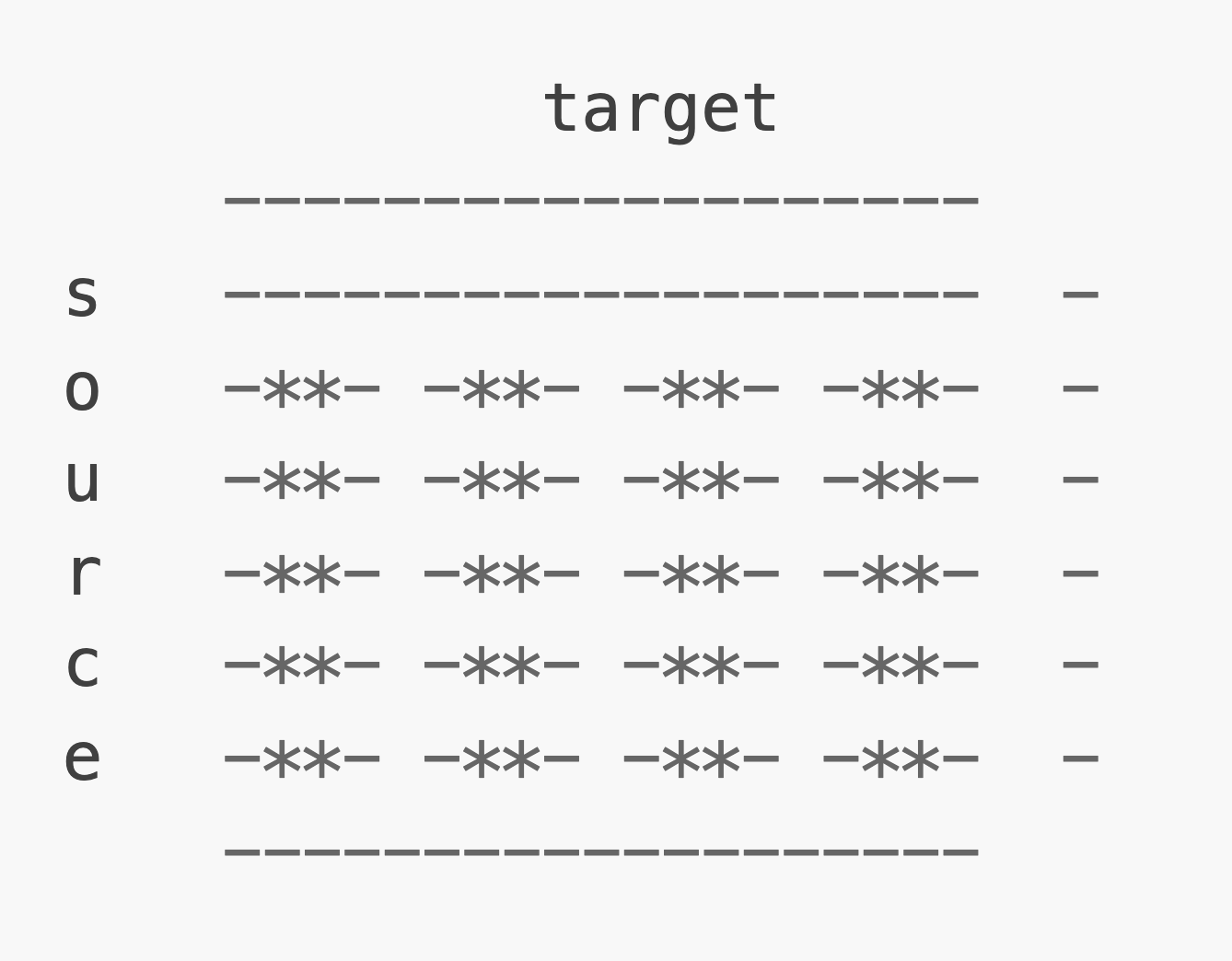}
\caption{Illustration of global quantization groups. All weights are considered together for scaling and quantizing, with distinct handling for output weights.}
\label{fig:global_quantization}
\end{figure}

\subsubsection{Channel Quantization}
In contrast, the channel quantization approach scales and quantizes parameters on a per-channel basis. This means that all input weights leading to a single target neuron are quantized together, allowing for more precise control over the weight scaling based on the specific needs of each neuron. This method is particularly useful when different neurons in the network have varying sensitivity to weight adjustments.

\begin{figure}[htbp]
\centering
\includegraphics[width=0.4\textwidth]{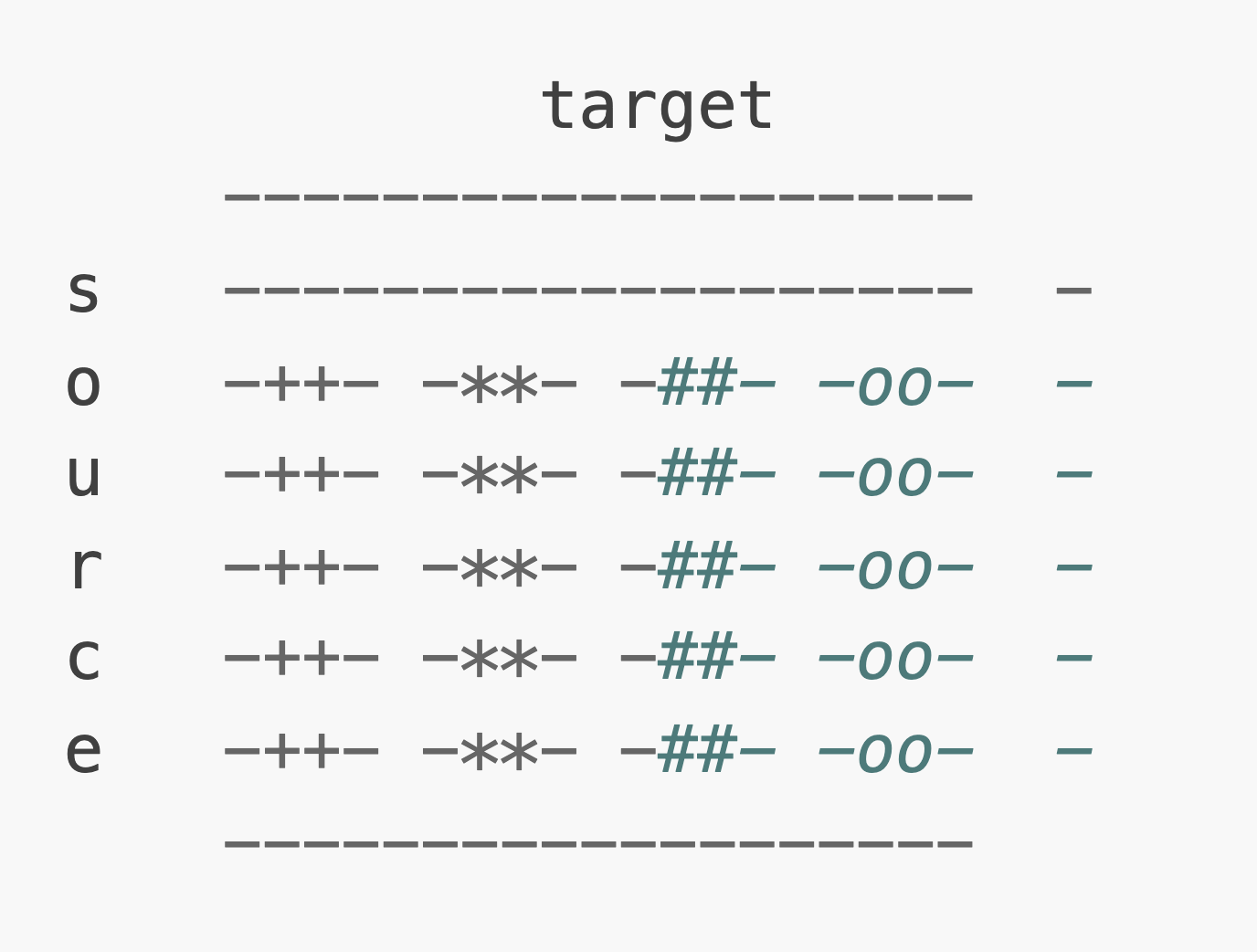}
\caption{Illustration of channel quantization groups. Weights leading to individual neurons are considered together, enabling finer adjustments for each channel.}
\label{fig:channel_quantization}
\end{figure}

% Both quantization methods ensure that after conversion, the resulting integer values appropriately represent the original floating-point numbers. This process is critical for deploying neural networks on hardware like the Xylo chip, where digital precision and memory constraints are paramount.

\subsection{Hardware Configuration and Deployment}
Once the network graph has been mapped to the Xylo hardware specifications using the \texttt{mapper()} function, the next crucial steps involve converting this specification into a hardware configuration and deploying it to the Xylo Hardware Development Kit (HDK).

The network specification is first converted into a hardware configuration object suitable for the Xylo HDK. This conversion is achieved using the \texttt{config\_from\_specification()} function. This function not only generates the configuration but also validates it to ensure that it meets all necessary hardware constraints. If the configuration fails to validate, a detailed message is provided, indicating the issues that need to be resolved. This ensures that only configurations that fully comply with the Xylo hardware specifications are considered for deployment.

After a valid hardware configuration is obtained, it is deployed to the Xylo HDK using the \texttt{XyloSamna()} function. This step actualizes the network on the hardware, allowing for real-time operation and testing. During deployment, specific parameters such as the timestep (\texttt{dt}) of the model are set, tailoring the deployment to match the operational timing requirements of the Xylo HDK. Successful deployment is confirmed by a response from the HDK, which typically includes details about the configuration's structure, such as its shape and dimensions, ensuring the model's operational integrity on the hardware.

% These steps collectively transform a theoretical neural network model into a practical, operational hardware setup on the Xylo platform, moving from digital simulation to tangible execution. This process not only highlights the integration of
\subsection{Network Evolution on the Xylo HDK}
After deploying the network configuration to the Xylo processor, inference can be performed in the integer-logic SNN processor core, and compared with the floating-point LIF simulation provided by Rockpool.

A Poisson input can be used, and transmitted to the Xylo processor to evolve the state of the SNN. During this evolution, all internal states such as membrane potentials, synaptic currents, and spiking events are recorded. It's important to note that while recording these internal states assists with debugging the network's functionality, it can reduce the speed of evolution, potentially below real-time operation. This trade-off is essential for detailed analysis but can be adjusted based on specific experimental needs.

Following the evolution process, the recorded data are visualized to analyze the network's performance. This visualization helps in identifying patterns, understanding the network's behavior, and making necessary adjustments to optimize performance.

\begin{figure}[htbp]
\centering
\includegraphics[width=0.8\textwidth]{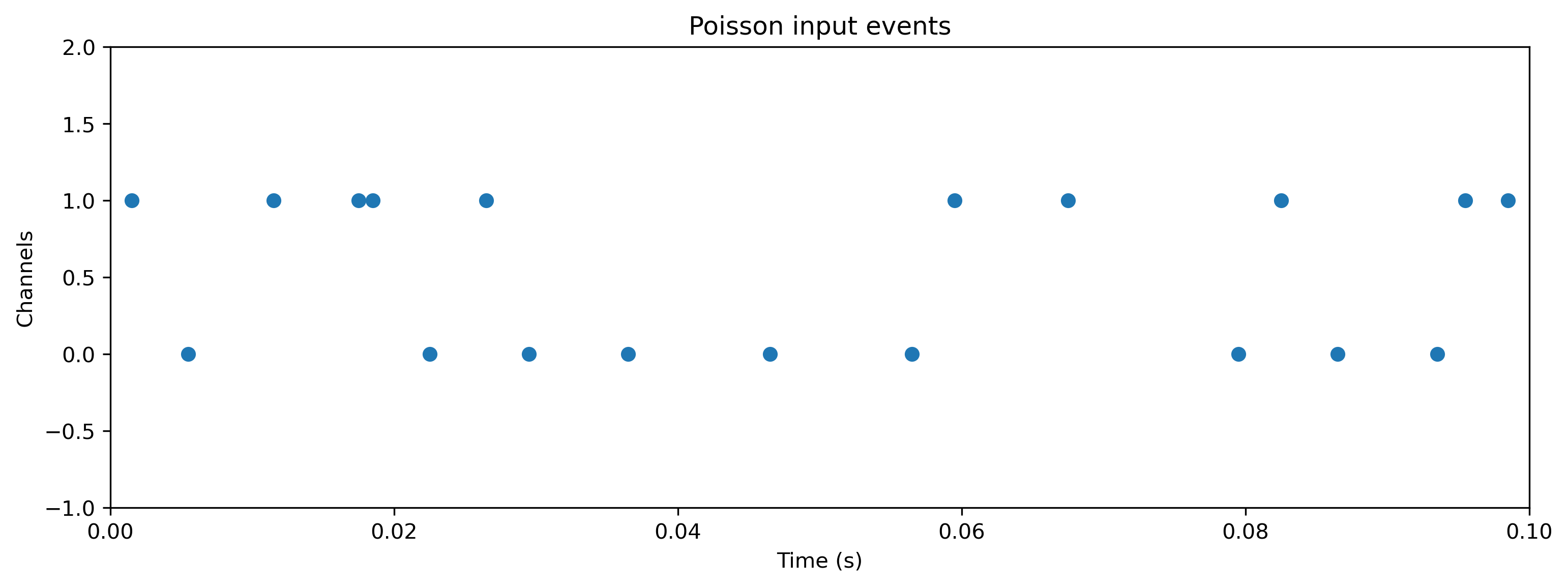}
\caption{Visualization of the generated Poisson spike train used as input for network stimulation.}
\label{fig:spike_train_input}
\end{figure}

\begin{figure}[htbp]
\centering
\includegraphics[width=1\textwidth]{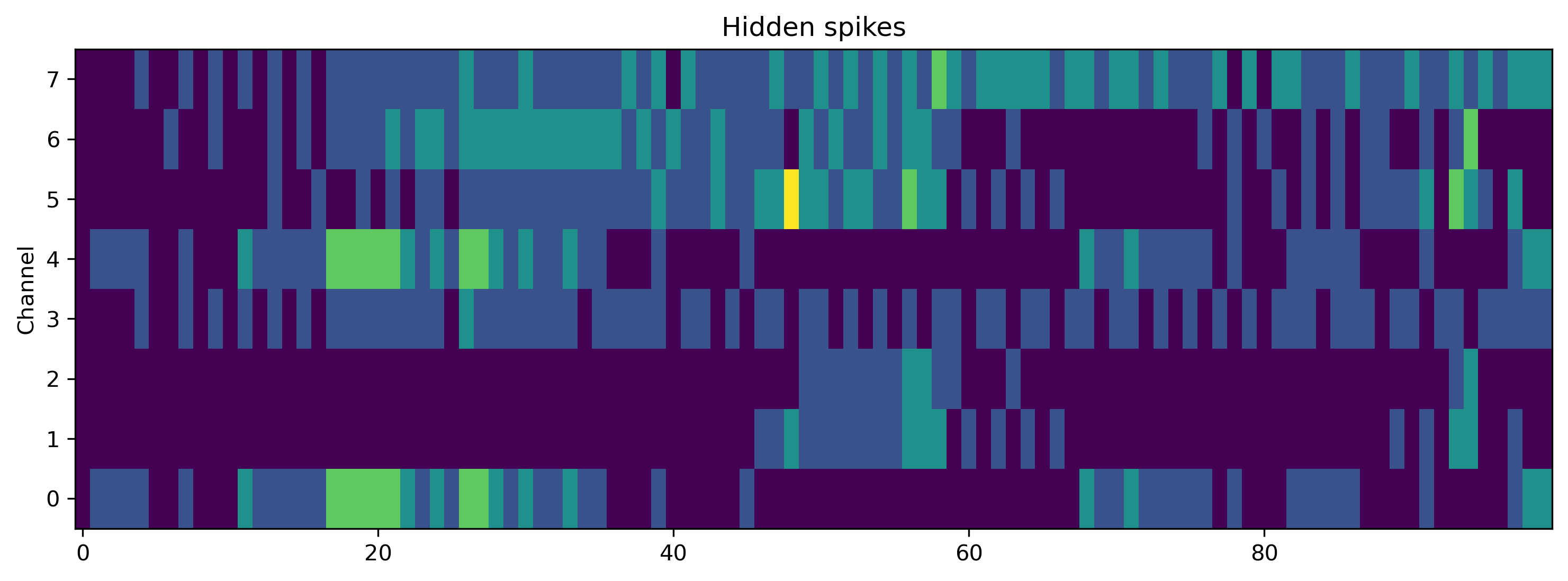}
\caption{Heatmap of network activity showing the spiking responses across different channels.}
\label{fig:network_activity}
\end{figure}

\begin{figure}[htbp]
\centering
\includegraphics[width=1\textwidth]{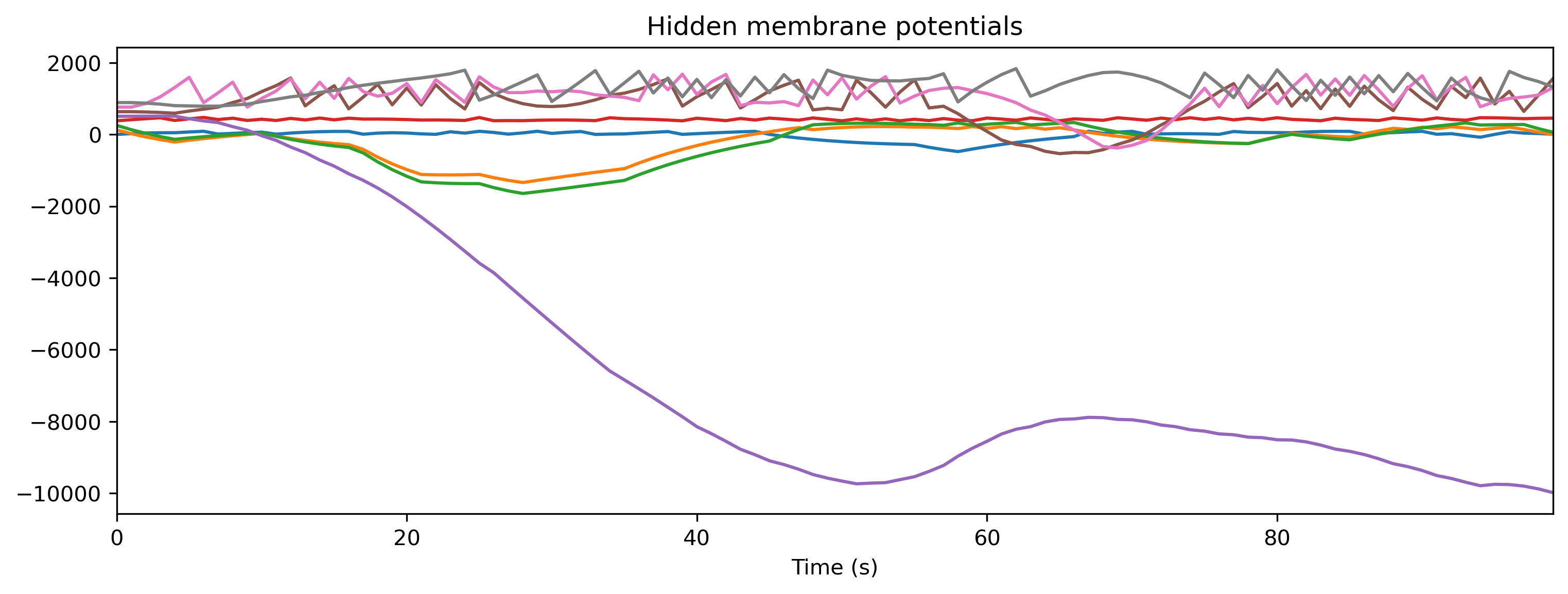}
\caption{Graph of membrane potentials over time, illustrating the dynamic response of the network to the Poisson input.}
\label{fig:membrane_potentials}
\end{figure}

\begin{figure}[htbp]
\centering
\includegraphics[width=1\textwidth]{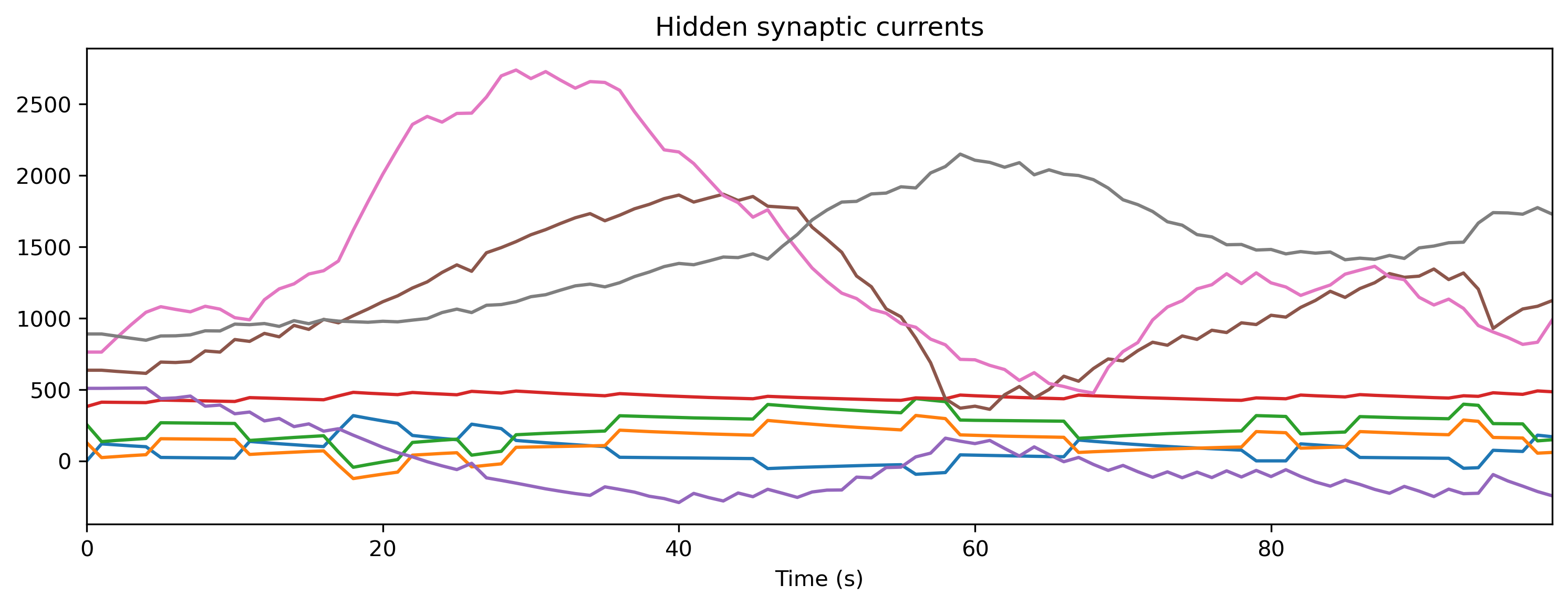}
\caption{Time-series plot of synaptic currents, providing insight into the synaptic integration within the network.}
\label{fig:synaptic_currents}
\end{figure}

% These visualizations play a crucial role in both qualitative and quantitative assessments of the neuromorphic system, providing comprehensive insights into its computational capabilities and responsiveness to stimuli.

% This evolution phase is fundamental as it bridges the gap between theoretical network design and practical application, providing a platform to test and refine the network in conditions that mimic real-world operations.

\section{Results}\label{sec5}
The final step in validating the performance of the deployed neural network on the Xylo HDK involves comparing the outputs generated by the hardware with those produced by XyloSim, a bit-precise simulator of the Xylo architecture. This comparison is crucial to demonstrate that the simulator can accurately reflect the hardware's behavior, providing a reliable platform for pre-deployment testing and verification.

The network was evolved using XyloSim with inputs identical to those used in the hardware deployment. Both platforms recorded various internal states such as membrane potentials, synaptic currents, and spiking events. The analysis of these recordings indicates a perfect match between the simulated data and the hardware output, showing that XyloSim accurately mirrors the real-time dynamics observed in the Xylo HDK.

Comparative visualizations of the outputs from both the simulator and the hardware provide a clear indication of their alignment. The following figures illustrate the results from the simulator, highlighting the close match in the spiking activity and synaptic behaviors recorded during the network evolution.

\begin{figure}[htbp]
\centering
\includegraphics[width=1\textwidth]{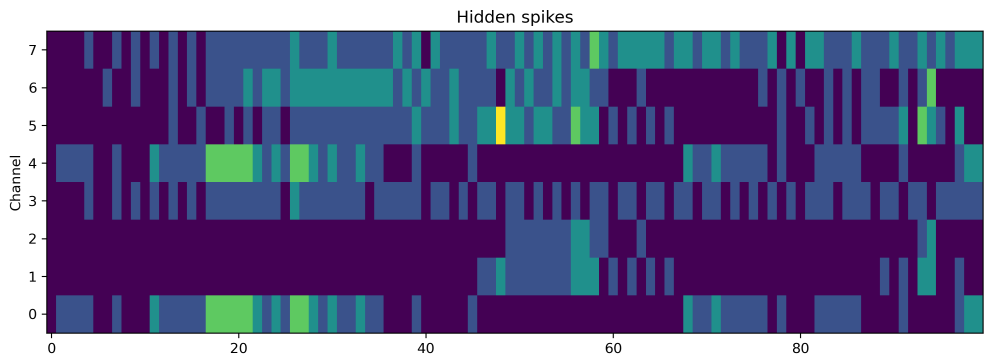}
\caption{Comparison of spike train outputs from the XyloSim and Xylo HDK, demonstrating the simulator's accuracy in reproducing the hardware's spiking patterns.}
\label{fig:comparison_spike_train}
\end{figure}

\begin{figure}[htbp]
\centering
\includegraphics[width=1\textwidth]{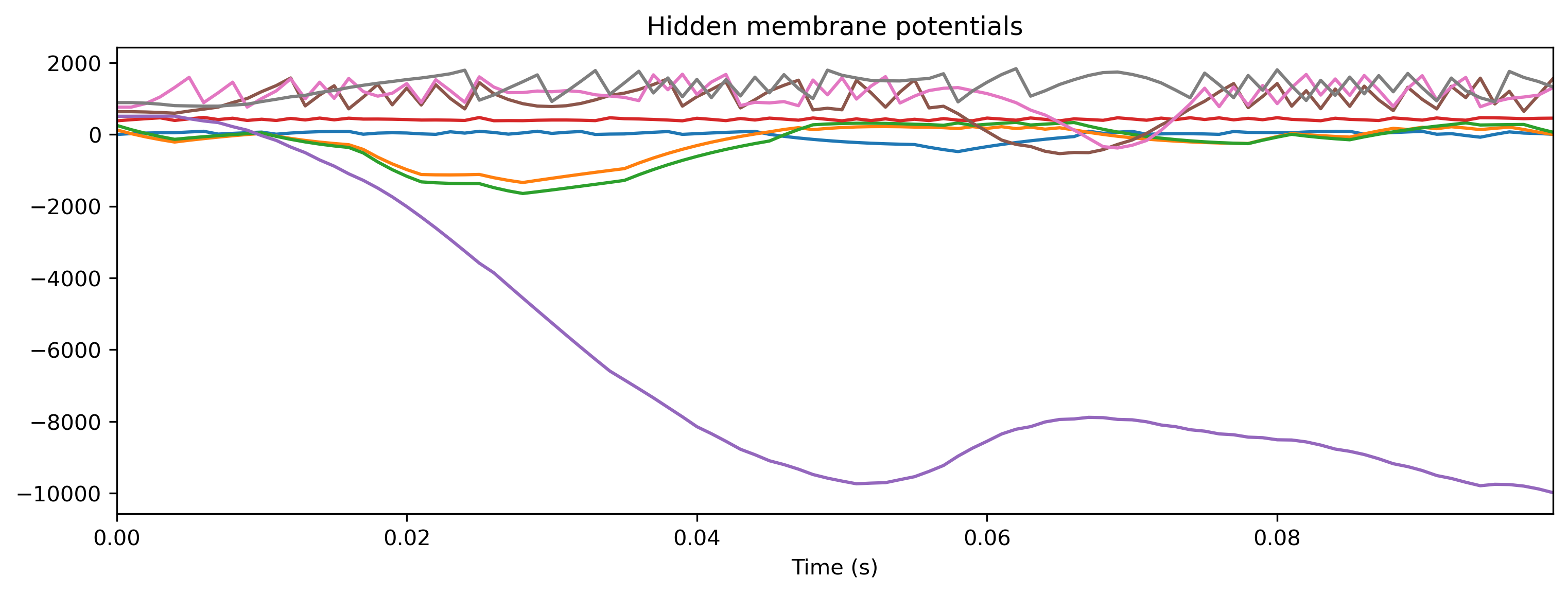}
\caption{Membrane potential dynamics as recorded from both the simulator and the hardware, showing their consistent behavior.}
\label{fig:comparison_membrane_potentials}
\end{figure}

\begin{figure}[htbp]
\centering
\includegraphics[width=1\textwidth]{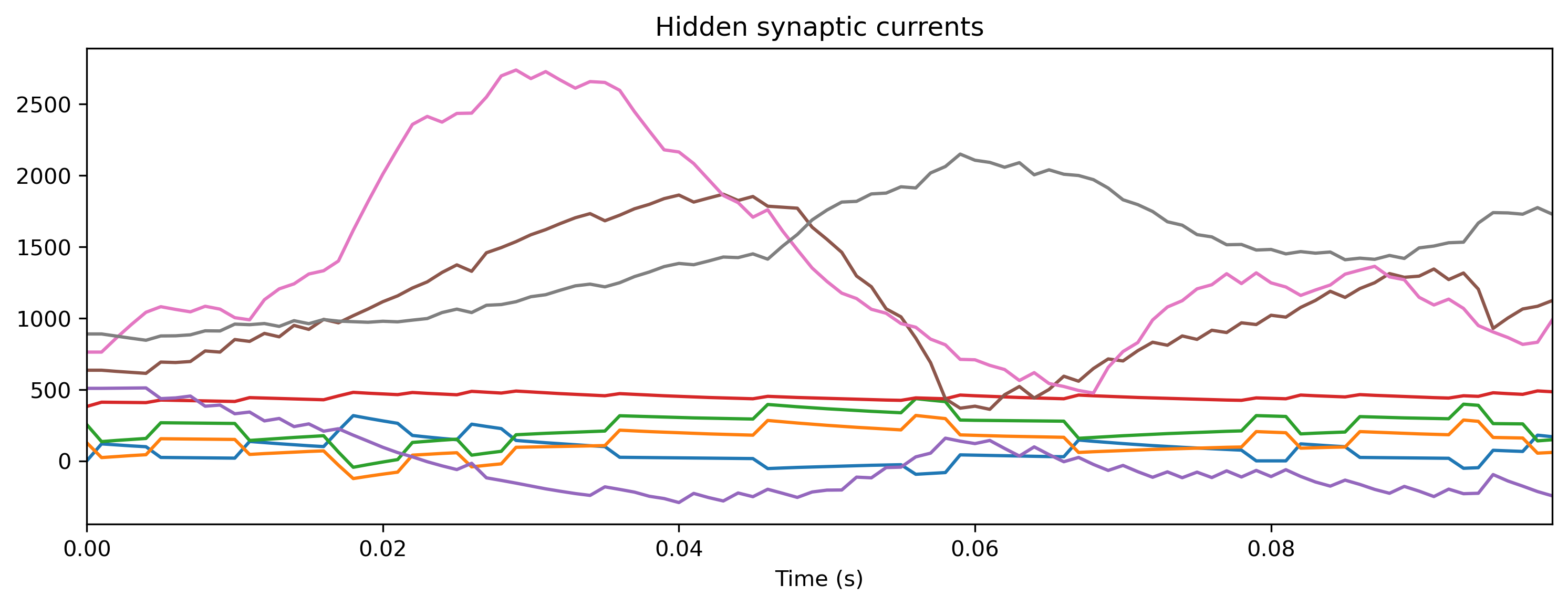}
\caption{Synaptic current profiles from the simulator and hardware, illustrating the detailed agreement between the two platforms.}
\label{fig:comparison_synaptic_currents}
\end{figure}

% The results confirm that the XyloSim provides a reliable and accurate representation of the Xylo HDK's operational characteristics without the need for physical hardware, facilitating earlier and more efficient testing phases. This capability is essential for optimizing neural network configurations before actual deployment, ensuring both effectiveness and efficiency in real-world applications.
% These findings underscore the utility of using a high-fidelity simulator like XyloSim to pre-validate the configurations and dynamic responses of neuromorphic hardware, significantly streamlining the development and deployment process.

\section{Conclusion}
The Rockpool to Xylo Hardware Development Kit (HDK) deployment pipeline marks a significant advancement in neuromorphic computing. It demonstrates a streamlined transition from conceptual neural network design to physical implementation, highlighting the efficacy of Rockpool in conjunction with the Xylo platform.

This pipeline methodically transforms theoretical models into deployable hardware configurations, encompassing network building, hardware specification mapping, and deployment. These steps ensure consistent and efficient final implementations, enhancing computational resource optimization and network performance.

Moreover, the pipeline's structured approach supports extensibility, replicability and scalability, making it adaptable for various applications in neuromorphic computing. It significantly advances the field, making sophisticated neural network applications more practical and accessible for research and industrial uses. This paves the way for future innovations in artificial intelligence and computational neuroscience.


\bibliography{sn-bibliography}% common bib file


\begin{thebibliography}{3}
\ifx \bisbn   \undefined \def \bisbn  #1{ISBN #1}\fi
\ifx \binits  \undefined \def \binits#1{#1}\fi
\ifx \bauthor  \undefined \def \bauthor#1{#1}\fi
\ifx \batitle  \undefined \def \batitle#1{#1}\fi
\ifx \bjtitle  \undefined \def \bjtitle#1{#1}\fi
\ifx \bvolume  \undefined \def \bvolume#1{\textbf{#1}}\fi
\ifx \byear  \undefined \def \byear#1{#1}\fi
\ifx \bissue  \undefined \def \bissue#1{#1}\fi
\ifx \bfpage  \undefined \def \bfpage#1{#1}\fi
\ifx \blpage  \undefined \def \blpage #1{#1}\fi
\ifx \burl  \undefined \def \burl#1{\textsf{#1}}\fi
\ifx \doiurl  \undefined \def \doiurl#1{\url{https://doi.org/#1}}\fi
\ifx \betal  \undefined \def \betal{\textit{et al.}}\fi
\ifx \binstitute  \undefined \def \binstitute#1{#1}\fi
\ifx \binstitutionaled  \undefined \def \binstitutionaled#1{#1}\fi
\ifx \bctitle  \undefined \def \bctitle#1{#1}\fi
\ifx \beditor  \undefined \def \beditor#1{#1}\fi
\ifx \bpublisher  \undefined \def \bpublisher#1{#1}\fi
\ifx \bbtitle  \undefined \def \bbtitle#1{#1}\fi
\ifx \bedition  \undefined \def \bedition#1{#1}\fi
\ifx \bseriesno  \undefined \def \bseriesno#1{#1}\fi
\ifx \blocation  \undefined \def \blocation#1{#1}\fi
\ifx \bsertitle  \undefined \def \bsertitle#1{#1}\fi
\ifx \bsnm \undefined \def \bsnm#1{#1}\fi
\ifx \bsuffix \undefined \def \bsuffix#1{#1}\fi
\ifx \bparticle \undefined \def \bparticle#1{#1}\fi
\ifx \barticle \undefined \def \barticle#1{#1}\fi
\bibcommenthead
\ifx \bconfdate \undefined \def \bconfdate #1{#1}\fi
\ifx \botherref \undefined \def \botherref #1{#1}\fi
\ifx \url \undefined \def \url#1{\textsf{#1}}\fi
\ifx \bchapter \undefined \def \bchapter#1{#1}\fi
\ifx \bbook \undefined \def \bbook#1{#1}\fi
\ifx \bcomment \undefined \def \bcomment#1{#1}\fi
\ifx \oauthor \undefined \def \oauthor#1{#1}\fi
\ifx \citeauthoryear \undefined \def \citeauthoryear#1{#1}\fi
\ifx \endbibitem  \undefined \def \endbibitem {}\fi
\ifx \bconflocation  \undefined \def \bconflocation#1{#1}\fi
\ifx \arxivurl  \undefined \def \arxivurl#1{\textsf{#1}}\fi
\csname PreBibitemsHook\endcsname

%%% 1
\bibitem[\protect\citeauthoryear{Muir et~al.}{2019}]{rockpool}
\begin{botherref}
\oauthor{\bsnm{Muir}, \binits{D.R.}},
\oauthor{\bsnm{Bauer}, \binits{F.}},
\oauthor{\bsnm{Weidel}, \binits{P.}}:
Rockpool Documentaton.
Zenodo
(2019).
\doiurl{10.5281/zenodo.3773845} .
\url{https://doi.org/10.5281/zenodo.3773845}
\end{botherref}
\endbibitem

%%% 2
\bibitem[\protect\citeauthoryear{Bos and Muir}{2023}]{bos2023sub}
\begin{bchapter}
\bauthor{\bsnm{Bos}, \binits{H.}},
\bauthor{\bsnm{Muir}, \binits{D.}}:
\bctitle{Sub-{mW} {Neuromorphic} {SNN} audio processing applications with {Rockpool} and {Xylo}}.
In: \bbtitle{Embedded Artificial Intelligence},
pp. \bfpage{69}--\blpage{78}.
\bpublisher{River Publishers}, \blocation{???}
(\byear{2023})
\end{bchapter}
\endbibitem

%%% 3
\bibitem[\protect\citeauthoryear{Bos and Muir}{2024}]{bos2024micro}
\begin{botherref}
\oauthor{\bsnm{Bos}, \binits{H.}},
\oauthor{\bsnm{Muir}, \binits{D.R.}}:
Micro-power spoken keyword spotting on {Xylo Audio 2}
(2024).
\url{https://arxiv.org/abs/2406.15112}
\end{botherref}
\endbibitem

\end{thebibliography}

\end{document}